\documentclass[letterpaper, 10 pt, conference]{IEEEtran} 

\usepackage{graphicx}
\usepackage{caption}
\graphicspath{{./}}
\usepackage[caption=false]{subfig}
\usepackage{placeins}
\usepackage{booktabs}

\usepackage[english]{babel}

\usepackage{soul, color}
\usepackage[hidelinks]{hyperref}

\IEEEoverridecommandlockouts     
 
\title{\LARGE \bf
Netherlands Dataset: A New Public Dataset for Machine Learning in Seismic Interpretation
}

\author{Reinaldo Mozart Silva$^{1}$, Lais Baroni$^{1}$, Rodrigo S. Ferreira$^{1}$, Daniel Civitarese$^{1}$ \\ Daniela Szwarcman$^{1}$, Emilio Vital Brazil$^{1}$
\thanks{$^{1}$Reinaldo Mozart Silva, Lais Baroni, Rodrigo S. Ferreira, Daniel Civitarese, Daniela Szwarcman and Emilio Vital Brazil are with IBM Research,
        Av. Pasteur 138/146, 22290-240, Botafogo, Rio de Janeiro, Brazil
        {\tt\small \{rmozart, lbaroni, rosife, sallesd, daniszw, evital\} at br.ibm.com}}
}

\begin{document}

\maketitle

\thispagestyle{empty}
\pagestyle{empty}

\begin{abstract}

Machine learning and, more specifically, deep learning algorithms have seen a remarkable growth in their popularity and usefulness in the last years. This is arguably due to three main factors: powerful computers, new techniques to train deeper networks and larger datasets. Although the first two are readily available in modern computers and ML libraries, the last one remains a challenge for many domains.
It is a fact that big data is a reality in almost all fields nowadays, and geosciences are not an exception. However, to achieve the success of general-purpose applications such as ImageNet -- for which there are +14 million labeled images for 1000 target classes -- we not only need more data, we need more high-quality labeled data.
When it comes to the Oil \& Gas industry, confidentiality issues hamper even more the sharing of datasets.
In this work, we present the Netherlands interpretation dataset, a contribution to the development of machine learning in seismic interpretation. The Netherlands F3 dataset acquisition was carried out in the North Sea, Netherlands offshore. The data is publicly available and contains pos-stack data, 8 horizons and well logs of 4 wells. For the purposes of our machine learning tasks, the original dataset was reinterpreted, generating 9 horizons separating different seismic facies intervals. The interpreted horizons were used to generate $\sim$190,000 labeled images for inlines and crosslines.
Finally, we present two deep learning applications in which the proposed dataset was employed and produced compelling results.

\end{abstract}

\section{INTRODUCTION}

Seismic interpretation plays a very important role in the exploration and production stream, being indispensable for geoscientists to delevelop oil and gas prospects. Being able to accurately describe the geology of the subsurface is a critical factor for the successful exploitation of known hydrocarbon accumulations \cite{herron2011first}.

However, seismic interpretation is a human-intensive and time-consuming task. The reflection seismic method only provides an indirect measurement of the subsurface geology, which is often noisy and limited in its spatial and temporal resolutions.
Thus, reconstructing the geologic story of a seismic survey still poses a tremendous challenge, despite the technological advances in data acquisition, processing and software applications \cite{herron2011first}. Tight deadlines and the increasing size of datasets are also complicating factors.




Other domains facing similar challenges are using neural networks and machine/deep learning techniques with great success to support tasks that deal with high volumes of data and are considered human-centered \cite{Badrinarayanan2015, Shelhamer, Redmon_2016_CVPR, NIPS2015_5638}.
Nevertheless, these methods require training datasets with a rather large amount of data \cite{goodfellow2016deep}.

The geoscience community, and more specifically the oil and gas industry, are already dealing with big data, but to fully harness the power of deep learning techniques, we need high-quality labeled datasets. This was one of the main factors that allowed general-purpose deep learning applications such as MNIST \cite{lecun2010} ($\sim$60k images), PASCAL-VOC \cite{everingham2010} ($\sim$40k images), MS-COCO \cite{lin2014} ($\sim$330k images), and ImageNet \cite{deng2009} ($\sim$14 million images) to be as successful as they are today. These applications comprise tasks like object detection, segmentation and classification, which have direct parallels in common tasks in the seismic interpretation workflow.


In this work, we present a contribution to the development of machine learning in seismic interpretation, which will be made publicly available. The Netherlands interpretation dataset consists of 9 horizons and $\sim$190{,}000 labeled seismic images derived from the Netherlands F3 seismic data \cite{osr}, already in the public domain.
The seismic sections were interpreted based on their seismic facies. Previous versions of the proposed dataset have already been used in some works such as \cite{chevitarese2018ijcnn, chevitarese2018methodology}, two of which will be discussed in Section \ref{sec:experiments}.

This paper is organized as follows: next section describes the regional geology of the Netherlands F3 survey. Sections \ref{sec:seismic_data} and \ref{sec:seismic_interpretation} present the Netherlands F3 seismic dataset and discuss the interpretation procedure. In Section \ref{sec:dataset} we present the proposed dataset and detail its main characteristics. Finally, in Sections \ref{sec:experiments} and \ref{sec:conclusions} we briefly discuss two deep learning applications in which the proposed dataset was employed and present our final remarks. 

\section{GEOLOGICAL SETTINGS}
\label{sec:geological_settings}

\begin{figure}
    \centering
    \includegraphics[width=0.5\textwidth]{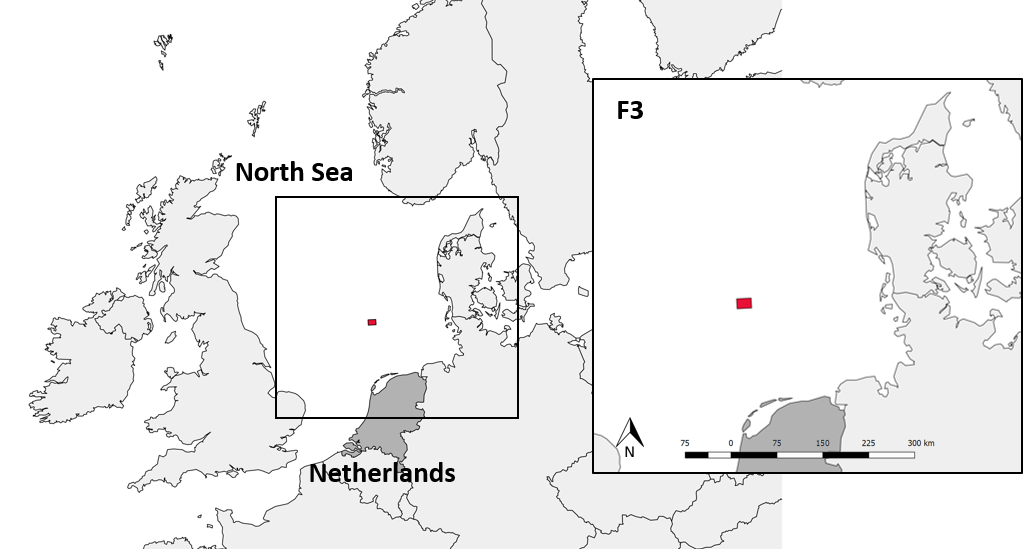}
    \caption{Location of the F3 3D survey in the North Sea, Netherlands offshore.}
    \label{fig:location} 
\end{figure}

The Netherlands F3 dataset is a seismic survey of approximately 384km\textsuperscript{2} in the Dutch offshore portion of the Central Graben basin, roughly situated at 180km from the Netherlands shore (Figure \ref{fig:location}) \cite{lucas2014caracterizaccao}.

The Central Graben Basin is a result of the North Atlantic opening and posterior division of the supercontinent Pangea. This main region is characterized by a triple rift system -- Central Graben, Viking Graben, and Moray Firth Basins \cite{ziegler1989evolution}.  

The formation of the Central Graben Basin transpired during the Mesozoic, after a triple rifting phase due to the three stages of the Atlantic Ocean opening \cite{alves2011modelaccao}. The first phase of subsidence is thermal, the second phase, during the Triassic, is represented by a rifting tectonic, ending with several extensional regimes during the Kimmeridgian \cite{schroot2003expressions}. 

As for the lithostratigraphic configuration, the Central Graben Basin is constituted of 9 main groups from the Carboniferous to the Cenozoic. According to \cite{duin2006subsurface}, the Carboniferous Group sediments -- the oldest stratigraphic record from the Central Graben Basin -- have limited information, available only from a few wells. This group is mainly composed of black limestones and general clastic rocks, reaching a thickness of more than 4{,}000m in some areas \cite{tno2004}. 

The following stratigraphic groups -- the Lower and Upper Rotliegend Group -- were deposited during the Early and Middle Permian. These deposits are mainly constituted of volcanic rocks and fluvio-lacustrine sediments, on the lower portions, and by fluvial, eolian, and sabkha sediments on the upper regions of the group, reaching a maximum thickness of 900m \cite{duin2006subsurface}. 

The Zechstein Group, the next lithostratigraphic portion, is composed by carbonate and evaporite rocks. In the Central Graben Basin, this group is characterized by several salt structures, reaching a thickness of 1{,}300m \cite{duin2006subsurface}. Moreover, in areas with intense tectonism, the upper layers are affected by the halokinesis caused by the evaporites of the Zechstein Group.

The subsequent group is the Germanic Trias Group, being deposited during the Triassic period, reaching a thickness of 1{,}800m \cite{duin2006subsurface}. The Lower portion of this group is mainly composed of red shale rocks and siltstone interbedded with sands \cite{wong1989late}, occurring during the Early Triassic. The Upper portion of the Germanic Trias Group, deposited during the Middle and Upper Triassic, is constituted of anhydrous evaporites \cite{wong1989late}. 

The following lithostratigraphic unit is the Altena Group, deposited during the Rhaetian untill the Callovian -- Middle Jurassic. Thick marine shales mainly represent this group, reaching a thickness of 1{,}600m \cite{duin2006subsurface}. 

The Late Jurassic deposits are represented by the Schieland, Scruff and Niedersachsen Groups. The thickness of this sediments, primarily lacustrine and clastic, had an intense influence of erosion and inversion, rarely reaching magnitude over 1{,}000m \cite{duin2006subsurface}. 

The deposits from the Early Cretaceous -- the Rijnland Group -- have no more than 1{,}000m thickness in specific regions due to intense erosion and inversion \cite{duin2006subsurface}. The sediments of this group are predominantly siliciclastic and suffer intense halokinesis on the Central Graben region.

The Chalk Group, the following lithostratigraphic unit, correspond to the Late Cretaceous deposits. Its sediments have more than 1{,}800m of thickness, on specific regions, and are mainly comprised of chalk and argillites.  However, most of the record is absent in the Dutch Central Graben basin due to inversion in the region \cite{duin2006subsurface}.

The North Sea Supergroup -- Lower, Middle, and Upper -- represents the remain deposits from the Cenozoic Era. In the Central Graben Basin, this period is characterized by strong subsidence, and have the most significant presence on the F3-Netherland seismic record. The North Sea Supergroup is also affected by intense halokinesis from the Zechstein Group salt.

\section{SEISMIC DATA}
\label{sec:seismic_data}

The seismic data used for the generation of the proposed dataset is a public 3D seismic survey called Netherlands Offshore F3 Block which is available at the Open Seismic Repository \cite{osr}. The dataset consists of 384km\textsuperscript{2} of time migrated 3D seismic data, with 651 inlines and 951 crosslines, located at the North Sea, Netherlands offshore (\autoref{fig:location}).

The seismic data has a time range of 1{,}848ms, a sampling rate of 4ms and a bin size of 25m. Along with the 3D seismic data, the repository also provides 4 wells, F02–1, F03–2, F03–4, and F06–1 -- with some markers and geophysical logs --, 8 interpreted horizons, and some additional 2D and 3D seismic data with attributes and models.

\section{SEISMIC INTERPRETATION}
\label{sec:seismic_interpretation}

The Netherlands F3 seismic dataset was reinterpreted by two geoscientists using the software OpendTect \cite{opendetect}. Although other data are available in the repository, the interpretation was produced based only on the 3D data, disregarding the provided horizons since they sometimes comprehend more than one significant texture, what could hamper the performance of the machine learning algorithms.

\begin{figure}
    \centering
    \includegraphics[width=0.45\textwidth]{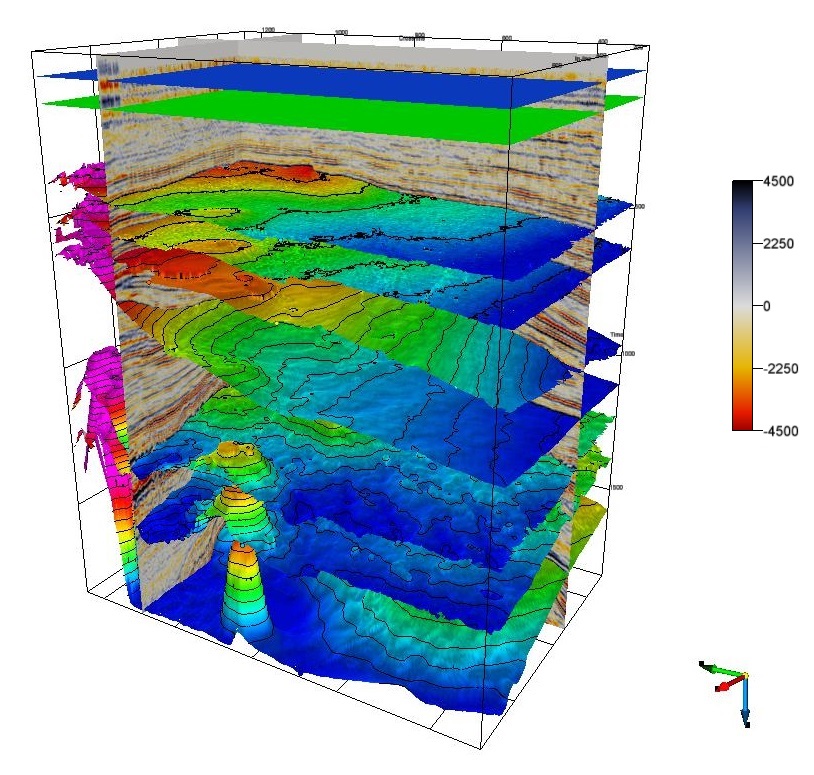}
    \caption{Nine interpreted horizons shown along with two seismic sections.}
    \label{fig:horizons} 

\vspace*{-3mm}

\end{figure}

Nine horizons were interpreted: H1, H2, H3, H4, H5, H6, H7, H8 and H9, sorted in descending order of geological age. \autoref{fig:horizons} shows the 9 interpreted horizons along with two seismic lines. It is noteworthy that interpreted horizons may not correspond to the top of formations or stratal interfaces since only pattern configurations were taken into consideration, focusing on the separation of different seismic facies intervals. One fault was also interpreted just to assist the horizons interpretation.

The analysis of seismic facies consists of the identification of seismic reflection parameters, based primarily on configuration patterns that indicate geological factors like lithology, stratification and depositional systems \cite{brown1980seismic}. In the following list, we describe the seismic facies of each horizon interval in terms of their amplitude and continuity.

 \begin{itemize}
   \item H1: this horizon limits the top of the Permian evaporite-rich succession of the Zechstein Group. The top reflectors feature stratified seismic facies while the lower ones present homogeneous seismic facies, where there are no apparent reflections.
   \item H2-H1: reflectors between H2 and H1 are continuous and have low amplitude. These characteristics are related to the neritic environment constituted mainly by argillaceous deposits.
   \item H3-H2: depositional environment and lithology of the sediments between H3 and H2 are essentially the same as H2, but this portion is marked by polygonal faults. Polygonal faults are common in argillaceous sediments and attribute discontinuous mound-shaped characteristic to reflectors.
   \item H4-H3: reflectors between H4 and H3 are predominantly subparallel and have varying amplitude.
   \item H5-H4: characterized by reflectors that has a prograding sigmoidal configuration of low energy and medium to low amplitude due to complex delta system deposited during the Late Miocene and Pliocene.
   \item H6-H5: package between H6 and H5 is made up mostly of parallel, high-amplitude reflectors due to Upper Cretaceous deposits of Chalk Group carbonates.
   \item H7-H6: comprises North Sea Supergroup, which has alternation of sandstones and claystones which give these facies the characteristics of semi-continuity and low amplitude.
   \item H8-H7: this interval still comprises sediments of the North Sea Supergroup and presents contorned to mound-shaped and low amplitude facies.
   \item H9-H8: facies between H9 and H8 are noisy, possibly because of acquisition noise or seismic processing failure.
 \end{itemize}

\section{NETHERLANDS INTERPRETATION DATASET}
\label{sec:dataset}

\begin{figure}
\centering
\includegraphics[width=.9\linewidth]{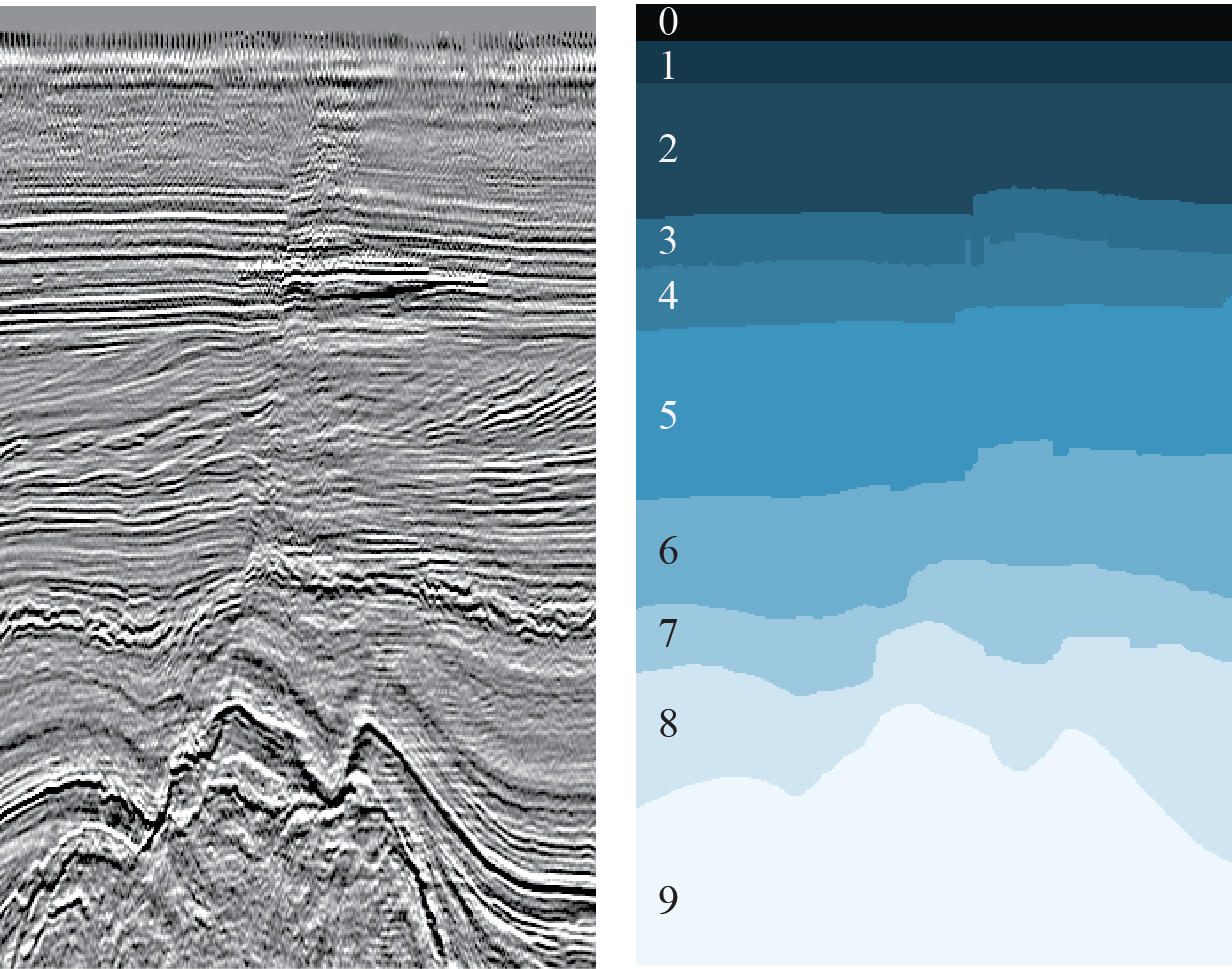}
\caption{Example of a cropped inline and its respective labels.}
\label{fig:dataset_example}
\end{figure}

The Netherlands interpretation dataset consists mainly of 9 interpreted horizons in XYZ format and 3{,}204 images, being 1{,}602 seismic lines in TIFF format and 1{,}602 labeled images in PNG format.
The labeled images were created by taking the intersection between the seismic lines and the horizon surfaces. Then, the pixels of each horizon interval were labeled from 0 to 9.
\autoref{fig:dataset_example} presents an inline section (cropped in the figure) and its respective labels.
In this paper, we present two applications: classification and semantic segmentation of seismic images. To make it easier for others to experiment with the dataset, we provide the image tiles used in the classification task within the package\footnote{The Netherlands interpretation dataset is available at: https://doi.org/10.5281/zenodo.1422787}.

To produce the classification dataset, we split the seismic images into tiles with 64$\times$25 pixels.
If the predominant class covers more than 70\% of the tile, it is associated with that class. Otherwise, the tile is discarded \cite{chevitarese2018methodology, chevitarese2018ijcnn}.
The process of creating tiles from a seismic image comprises the following steps:





\subsubsection{stretch the contrast and re-scale values between 0 and 255} this step removes extreme amplitudes and allows for a more compact representation of the dataset, e.g. using 8-bit integers.


\subsubsection{generate tiles from processed images} in this step we split the image in tiles and associate each tile with its predominant class.


\subsubsection{balance training dataset} there are many solutions to deal with imbalanced datasets. In this work, we decided to simply balance the number of samples per class, which makes the training process easier and allows us to use standard metrics such as accuracy.


The provided classification dataset includes 9{,}440 crossline and 9{,}472 inline seismic tiles \textit{per class} along with their respective labels. \autoref{tab:dataset} describes the files in the dataset.
We do not provide ready-to-use validation and test sets. Thus, users can create their own splits. A common practice is to separate 25\% of the dataset for test, and from the remaining training dataset, keep 25\% for validation.
The name of the tile image files contains the original inline/crossline number, so users can guarantee that the tiles in one seismic section go either in the training or test dataset. 

\begin{table}[]
\caption{Details of the Netherlands interpretation dataset}
\label{tab:dataset}
\centering
\begin{tabular}{@{}llrr@{}}
\toprule
File               & Format & \# Files & Total size (MB) \\ \midrule
H1-H9              & XYZ    & 9        & 225        \\
Seismic inlines    & TIF    & 651      & 1,150       \\
Seismic crosslines & TIF    & 951      & 1,150       \\
Labeled inlines    & PNG    & 651      & 2.6         \\
Labeled crosslines & PNG    & 951      & 2.9         \\
Seismic tiles (inlines) & PNG    & 94720   & 140       \\ 
Seismic labels (inlines) & JSON    & 1        & 2.7    \\ 
Seismic tiles (crosslines) & PNG    & 94400   & 141    \\ 
Seismic labels (crosslines) & JSON    & 1        & 2.9    \\ \bottomrule
\end{tabular}
\end{table}

\section{EXPERIMENTS}
\label{sec:experiments}

In this section, we present two deep learning applications that employ the Netherlands interpretation dataset: classification of rock layers (strata) and semantic segmentation of seismic images.

\subsection{Classification of Rock Layers}

The first application is the classification of different types of rock layers as presented in \cite{chevitarese2018ijcnn} and \cite{chevitarese2018methodology}. In both works, the authors trained deep neural networks that were able to successfully discriminate strata in the Netherlands F3 dataset. The main assumption is that one may distinguish different layers by their textural features as discussed in \cite{mattos2017assessing, chopra2006applications}. Hence, a model that can classify images based on their textural attributes could be used to classify distinct rock types.

In the experiments presented in \cite{chevitarese2018methodology}, the dataset preprocessing step included a sliding window mechanism to split the input image into tiles, in which each tile received the label of its predominant texture. These small images were then used as the actual input of the model that classifies each tile as one of the possible classes.

\begin{figure}
\centering
\includegraphics[width=.9\linewidth]{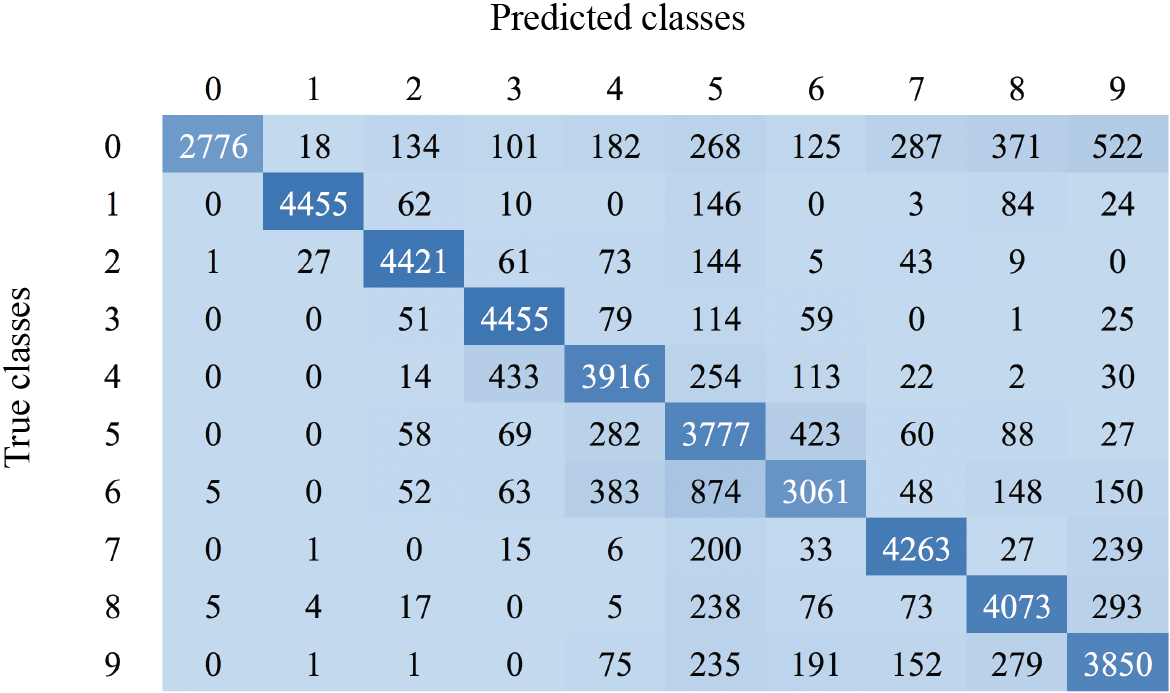}
\caption{Confusion matrix for rock layer classification using the proposed dataset.}
\label{fig:confusion_matrix}
\end{figure}

The authors in \cite{chevitarese2018methodology} tested multiple tile sizes, many numbers of examples per rock type, different percentages for texture predominance among other parameters, using a similar training dataset based on Netherlands F3. To create a baseline, we applied the same methodology on the proposed interpretation dataset. In this experiment, we used only 9 seismic sections randomly selected from the odd lines of the cube for training and validation. From these sections, we took 80\% for training and 20\% for validation. For testing, we used all the even lines which produced 4,784 tiles per class, after balancing. The final accuracy in the test dataset was 81.6\%. \autoref{fig:confusion_matrix} shows the final confusion matrix.

\subsection{Semantic Segmentation of Seismic Images}

The second application in which the Netherlands interpretation dataset was used is the semantic segmentation of seismic images. In \cite{chevitarese2018segmentation}, the authors selected a trained deep neural network from the rock layer classification task \cite{chevitarese2018methodology} to function as a pre-trained feature extractor. This transfer learning technique makes training faster and more accurate, as discussed in \cite{chevitarese2018segmentation, chevitarese2018transfer}.

Next, they removed the tail of the network that acts as a classifier and appended an upscale module to produce pixel-wise predictions based on the the main extracted features. Such a process has already been used in many other works \cite{ronneberger2015unet, long2015fully}. Finally, they trained the resulting model using the Netherlands interpretation dataset.

\begin{figure}
\centering
\includegraphics[width=.95\linewidth]{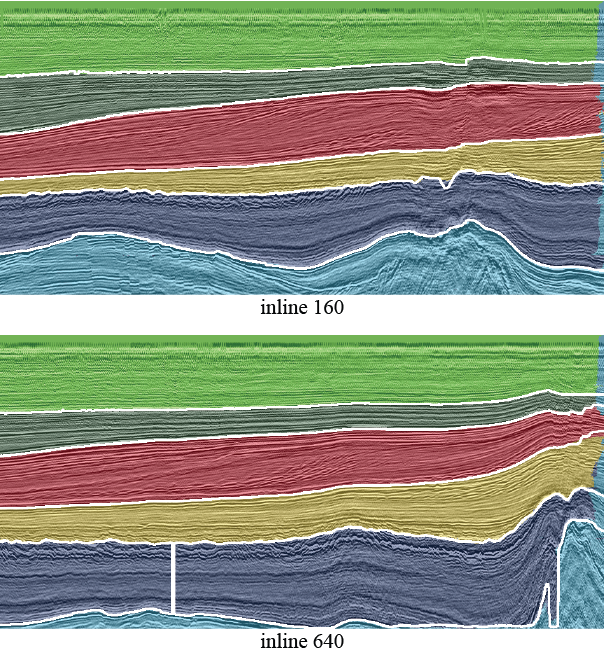}
\caption{Semantic segmentation of inlines 160 (top) and 640 (bottom) from the Netherlands F3 dataset. In the output, each pixel receives an overlaid color representing their class while the white lines represent the reference seismic horizons (ground truth). In that work, the authors used a subset of the horizons/classes present in the proposed Netherlands interpretation dataset.}
\label{fig:result}
\end{figure}

Similarly to the first application discussed, the authors in \cite{chevitarese2018segmentation} divided the input seismic section into small tiles. Additionally, they merged some layers to prevent thin layers from unbalancing the dataset. Next, they applied the network throughout the image to generate the final prediction. By doing this, they achieved more than 98\% of the mean Intersection over Union (IoU) metric. \autoref{fig:result} shows that the model produced segmentations very close to the actual interpretation (white lines) with very little discontinuity.

Although ready-to-use tile data are provided with the dataset, users can split the original seismic sections using different tile sizes, depending on the task at hand. For example, the result obtained in \cite{chevitarese2018segmentation} was produced using more than 50\% of the seismic lines for training. However, the authors have been able to achieve $\sim$90\% of mean IoU for semantic segmentation using only 9 training lines and a tile size of 120$\times$80 pixels for the same dataset.

\section{CONCLUSIONS}
\label{sec:conclusions}

In this work, we argued that seismic interpretation is a human-intensive and time-consuming task which is indispensable for the identification and exploitation of hydrocarbon accumulations. On the other hand, deep learning techniques have seen a remarkable growth while achieving impressive results in similar human-centered tasks. Many of them being general-purpose applications related to object detection, segmentation and classification, which have obvious parallels in the seismic interpretation workflow.

However, for the successful application of such techniques on seismic interpretation, we not only have to leverage the large amounts of data produced daily in oil and gas companies but also to create high-quality labeled data. With our dataset, we make a contribution that will allow geoscientists and machine learning practitioners working in the field to validate their models and compare their results.

In the experiments discussed in this work, the authors used the proposed dataset to train state-of-the-art deep learning models for rock layer classification and semantic segmentation of seismic images, obtaining high accuracies. Nevertheless, the Netherlands interpretation dataset could also be used to train and validated other machine learning techniques such as clustering, retrieval and transfer learning.

\bibliographystyle{ieeetr}
\bibliography{references}

\end{document}